\documentclass[runningheads]{llncs}
\usepackage{graphicx}
\usepackage{xcolor}
\usepackage{amsfonts}
\usepackage{bbding}
\begin{document}
\title{Knowledge distillation from multi-modal to mono-modal segmentation networks}

\titlerunning{KD-Net}
%
\author{
Minhao Hu\inst{1, 2\star} \and
Matthis Maillard\inst{2\star(}\Envelope\inst{)} \and
Ya Zhang\inst{1(}\Envelope\inst{)} \and
Tommaso Ciceri\inst{2} \and
Giammarco La Barbera \inst{2} \and
Isabelle Bloch \inst{2} \and
Pietro Gori \inst{2}}

%
\authorrunning{M.Hu et al.}
\institute{CMIC, Shanghai Jiao Tong University, Shanghai, China 
\and
LTCI, Télécom Paris, Institut Polytechnique de Paris, France
\email{matthis.maillard@telecom-paris.fr \\ ya\_zhang@sjtu.edu.cn}
}
\renewcommand{\thefootnote}{\fnsymbol{footnote}}
\footnotetext[1]{The two first authors contributed equally to this paper.}
\maketitle              
\begin{abstract}
The joint use of multiple imaging modalities for medical image segmentation has been widely studied in recent years. The fusion of information from different modalities has demonstrated to improve the segmentation accuracy, with respect to mono-modal segmentations, in several applications. However, acquiring multiple modalities is usually not possible in a clinical setting due to a limited number of physicians and scanners, and to limit costs and scan time. Most of the time, only one modality is acquired. In this paper, we propose KD-Net, a framework to transfer knowledge from a trained multi-modal network (teacher) to a mono-modal one (student). The proposed method is an adaptation of the generalized distillation framework where the student network is trained on a subset (1 modality) of the teacher's inputs (n modalities). We illustrate the effectiveness of the proposed framework in brain tumor segmentation with the BraTS 2018 dataset. Using different architectures, we show that the student network effectively learns from the teacher and always outperforms the baseline mono-modal network in terms of segmentation accuracy.

\end{abstract}

\section{Introduction}

Using multiple modalities to automatically segment medical images has become a common practice in several applications, such as brain tumor segmentation~\cite{brats_2015} or ischemic stroke lesion segmentation~\cite{maier_isles_2017}. Since different image modalities can accentuate and better describe different tissues, their fusion can improve the segmentation accuracy.  Although multi-modal models usually give the best results, it is often difficult to obtain multiple modalities in a clinical setting due to a limited number of physicians and scanners, and to limit costs and scan time. In many cases, especially for patients with pathologies or for emergency, only one modality is acquired.

Two main strategies have been proposed in the literature to deal with problems where multiple modalities are available at training time but some or most of them are missing at inference time.
The first one is to train a generative model to synthesize the missing modalities and then perform multi-modal segmentation. In \cite{van_tulder_why_2015}, the authors have shown that using a synthesized modality helps improving the accuracy of classification of brain tumors. Ben Cohen et al. \cite{ben_cohen_2018} generated PET images from CT scans to reduce the number of false positives in the detection of malignant lesions in livers. Generating a synthesized modality has also been shown to improve the quality of the segmentation of white matter hypointensities \cite{orbes-arteaga_simultaneous_2018}. The main drawback of this strategy is that it is computationally cumbersome, especially when many modalities are missing. In fact, one needs to train one generative network per missing modality in addition to a multi-modal segmentation network.

The second strategy consists in learning a modality-invariant feature space that encodes the multi-modal information during training, and that allows for all possible combinations of modalities during inference. Within this second strategy, Havaei et al. proposed HeMIS \cite{havaei_hemis:_2016}, a model that, for each modality, trains a different feature extractor. The first two moments of the feature maps are then computed and concatenated in the latent space from which a decoder is trained to predict the segmentation map.
Dorent et al. \cite{shen_hetero-modal_2019}, inspired by HeMIS, proposed U-HVED where they introduced skip-connections by considering intermediate layers, before each down-sampling step, as a feature map. This network outperformed HeMIS on BraTS 2018 dataset. In \cite{chen_robust_2019}, instead of fusing the layers by computing mean and variance, the authors learned a mapping function from the multiple feature maps to the latent space. They claimed that computing the moments to fuse the maps is not satisfactory since it makes each modality contribute equally to the final result which is inconsistent with the fact that each modality highlights different zones. They obtained better results than HeMIS on BraTS 2015 dataset.
This second strategy has good results only when one or two modalities are missing, however, when only one modality is available, it has worse results than a model trained on this specific modality. This kind of methods is therefore not suitable for a clinical setting where only one  modality is usually acquired, such as pre-operative neurosurgery or radiotherapy.

In this paper, in contrast to the previously presented methods, we propose a framework to \textit{transfer knowledge} from a multi-modal network to a mono-modal one. The proposed method is based on \textit{generalized knowledge distillation} \cite{lopez-paz_unifying_2016} which is a combination of distillation \cite{hinton_distilling_2015} and privileged information \cite{vapnik_2015}. \textit{Distillation} has originally been designed for classification problems to make a small network (Student) learn from an ensemble of networks or from a large network (Teacher). It has been applied to image segmentation in \cite{liu_structured_,xie_improving_2018} where the \textit{same input modalities} have been used for the Teacher network and the Student network. In \cite{xie_improving_2018}, the Student learns from the Teacher only thanks to a loss term between their outputs. In \cite{liu_structured_}, the authors also constrained the intermediate layers of the Student to be similar to the ones of the Teacher. With a different perspective, the framework of \textit{privileged information} was  designed to boost the performance of a Student model by learning from both the training data and a Teacher model with privileged and additional information. 
In generalized knowledge distillation, one uses distillation to extract useful knowledge from the privileged information of the Teacher \cite{lopez-paz_unifying_2016}. In our case, Teacher and Student have the same architecture (i.e. same number of parameters) but the Teacher can learn from multiple input modalities (additional information) whereas the Student from only one. The proposed framework is based on two encoder-decoder networks, which have demonstrated to work well in image segmentation \cite{isensee_2018}, one for the Student and one for the Teacher. Importantly, the proposed framework is generic since it works for different architectures of the encoder-decoder networks.
Each encoder summarizes its input space to a latent representation that captures important information for the segmentation. Since the Teacher and the Student process different inputs but aim at extracting the same information, we make the assumption that their first layers should be different, whereas the last layers and especially the latent representations (i.e. bottleneck) should be similar. By forcing the latent space of the Student to resemble the one of the Teacher, we make the hypothesis that the Student should learn from the additional information of the Teacher.
To the best of our knowledge, this is the first time that the generalized knowledge distillation strategy is adapted to guide the learning of a mono-modal student network using a multi-modal teacher network. 
We show the effectiveness of the proposed method using the BraTS 2018 dataset \cite{brats_2015} for brain tumor segmentation. 

\begin{figure}
\includegraphics[width=\textwidth]{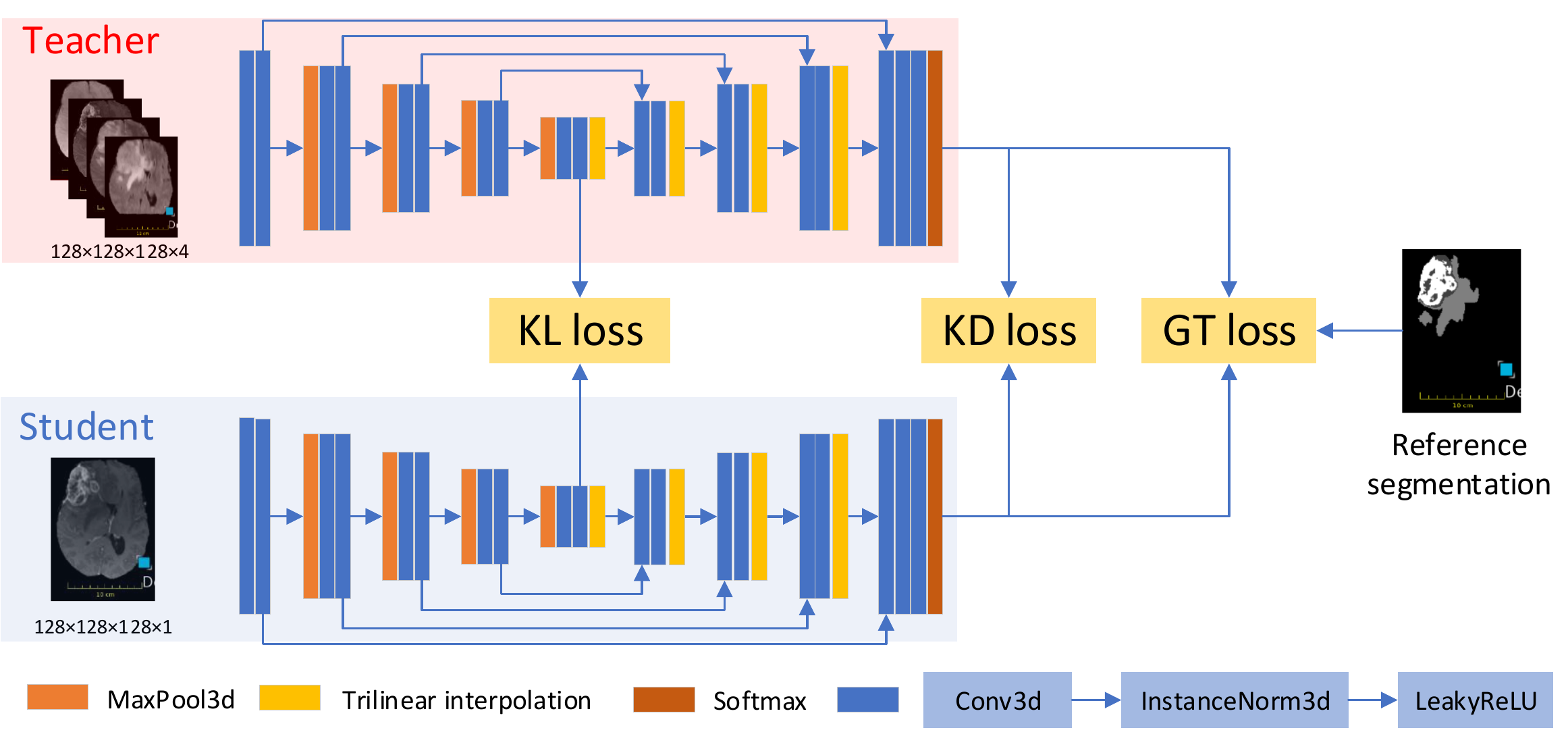}
\caption{Illustration of the proposed framework. Both Teacher and Student have the same architecture adapted from nnUNet~\cite{isensee_2018}. First, the Teacher is trained using only the reference segmentation (GT loss). Then, the student network is trained using all proposed losses: KL loss, KD loss and GT loss.} \label{fig1}
\end{figure}

The paper is organized as follows. First, we present the proposed framework, called KD-Net and illustrated in Figure~\ref{fig1}, and how the Student learns from the Teacher and the reference segmentation. Then, we present the implementation details and the results on the BraTS 2018 dataset \cite{brats_2015}. 

\section{KD-Net}

The goal of the proposed framework is to train a mono-modal segmentation network (Student) by leveraging the knowledge from a well-trained multi-modal segmentation network (Teacher). Except for the number of input channels, both networks have the same encoder-decoder architecture with skip connections. The multi-modal input $\mathrm{x^i}=\{x_n^i, {n=1...N}\}$ is the concatenation of the $N$ modalities for the $i^{th}$ sample of the dataset. Let $E_{t}$ and $D_{t}$ (resp. $E_{s}$ and $D_{s}$) denote the encoder and decoder parts of the Teacher (resp. Student). The Teacher network $f_{t}(\mathrm{x^i})= D_{t}\circ E_{t}(\mathrm{x^i})$ receives as input multiple modalities whereas the student network $f_{s}(x_k^i)=D_{s} \circ E_{s}(x_k^i)$ only one modality $x_k^i$, $k$ being a fixed integer between $1$ and $N$. 

We first train the Teacher, using only the reference segmentation as target. Then, we train the Student using three different losses: the knowledge distillation term, the dissimilarity between the latent spaces, and the reference segmentation loss. Note that the weights of the Teacher are frozen during the training of the Student and the error of the Student is not back-propagated to the Teacher.

The first two terms allow the Student to learn from the Teacher by using the soft prediction of the latter as target and by forcing the encoded information (i.e. bottleneck) of the Student to be similar to the one of the Teacher. The last term makes the predicted segmentation of the Student similar to the reference segmentation.

\subsection{Generalized knowledge distillation}\label{sec:KD}
  Following the strategy of generalized knowledge distillation \cite{lopez-paz_unifying_2016}, we transfer useful knowledge from the additional information of the Teacher to the Student using the soft label targets of the Teacher. These are computed as follows: 
    \begin{equation}\label{EQsoft}
  s_i = \sigma(f_t(\mathrm{x^i})/T)
  \end{equation}
    where $\sigma$ is the softmax function and $T$, the temperature parameter, is a strictly positive value. The parameter $T$ controls the softness of the target, and the higher it is, the softer the target. The idea of using soft targets is to uncover relations between classes that would be harder to detect with hard labels. The effectiveness of using a temperature parameter to soften the labels was demonstrated in \cite{hinton_distilling_2015}. 
  
  The knowledge distillation loss is defined as:
  \begin{equation}\label{EQKD}
  L_{KD} = \sum_i \left[(1-Dice(s_i, \sigma(f_s(x_k^i)))) + BCE(s_i^*,\sigma(f_s(x_k^i))\right]
  \end{equation}
    where $Dice$ is the Dice score, $BCE$ the binary cross-entropy measure and $s_i^*$ the binary prediction of the teacher. We need to binarize $s_i$ since the soft labels cannot be used in the binary cross-entropy.  
  The dice score ($Dice$) measures the similarity of the shape of two ensembles. Hence, it globally measures how the Teacher and Student's segmentation maps are close to each other. By contrast, the binary cross-entropy ($BCE$) is computed for each pixel independently and therefore it is a local measure. We use the combination of these two terms to globally and locally measure the distance between the Student prediction and the Teacher soft labels. 
  
  \subsection{Latent space}
  We speculate that Teacher and Student, having different inputs, should also encode differently the information in the first layers, the ones related to low-level image properties, such as color, texture and edges. By contrast, the deepest layers closer to the bottleneck, and related to higher level properties, should be more similar. Furthermore, we make the assumption that an encoder-decoder network encodes the information to correctly segment the input images in its latent space. Based on that, we propose to force the Student to learn from the additional information of the Teacher encoded in its bottleneck (and partially in the deepest layers) by making their latent representations as close as possible. To this end, we apply the Kullback-Leibler (KL) divergence as a loss function between the teacher and student's bottlenecks:
   \begin{equation}\label{EQKL}
     L_{KL}(p,q) = \sum_i\sum_j q_i(j) \log{\left(\frac{q_i(j)}{p_i(j)}\right)} 
  \end{equation}
     where $p_i$ (resp. $q_i$) are the flattened and normalized vector of the bottleneck $E_s(x^i_k)$ (resp $E_t(\mathrm{x}^i)$). Note that this function is not symmetric and we put the vectors in that order because we want the distribution of the Student's bottleneck to be similar to the one of the Teacher.
  
  \subsection{Objective function}
  We add a third term to the objective function to make the predicted segmentation as close as possible to the reference segmentation. It is the sum of the Dice loss ($Dice$) and the binary cross-entropy ($BCE$) for the same reasons as in Section~\ref{sec:KD}. We call it $L_{GT}$:
  \begin{equation}\label{EQGT}
  L_{GT} = \sum_i \left[(1-Dice(y_i, \sigma(f_s(x_k^i)))) + BCE(y_i,\sigma(f_s(x_k^i))\right].
   \end{equation}
  where $y_i$ denotes the reference segmentation of the $i^{th}$ sample in the dataset.
  
  The complete objective function is then: 
  \begin{equation}\label{EQobj}
  L = \lambda L_{KD} + (1-\lambda) L_{GT} + \alpha L_{KL}
  \end{equation}
   with $\lambda \in [0,1]$ and $\alpha \in \mathbb{R}^+$. The imitation parameter $\lambda$ balances the influence of the reference segmentation with the one of the Teacher's soft labels. The greater the $\lambda$ value, the greater the influence of the Teacher's soft labels. The $\alpha$ parameter is instead needed to balance the magnitude of the KL loss with respect to the other two losses.

\section{Results and Discussion}
\subsection{Dataset}
We evaluate the performance of the proposed framework on a publicly available dataset from the BraTS 2018 Challenge \cite{brats_2015}. It contains MR scans from 285 patients with four modalities: T1, T2, T1 contrasted-enhanced (T1ce) and Flair. The goal of the challenge is to segment three sub-regions of brain tumors: whole tumor (WT), tumor core (TC) and enhancing tumor (ET). We apply a central crop of size $128 \times 128 \times 128$ and a random flip along each axis for data augmentation. For each modality, only non-zero voxels have been normalized by subtracting the mean and dividing by standard deviation. Due to memory and time constraint, we subsample the images to the size $64 \times 64 \times 64$.

\subsection{Implementation details}

We adopt the encoder-decoder architecture described in Figure \ref{fig1}. Empirically, we found that the best parameters for the objective function are $\lambda = 0.75$, $T=5$ and $\alpha=10$. We used Adam optimizer for 500 epochs with a learning rate equal to 0.0001 that is multiplied by 0.2 when the validation loss has not decreased for 50 epochs. We run a three fold cross validation on the 285 training cases of BraTS 2018. The training of the baseline, the Teacher or the Student takes approximately 12 hours on a NVIDIA P100 GPU. 

\subsection{Results}
In our experiments, the Teacher uses all four modalities (T1, T2, T1ce and Flair concatenated) and the Student uses only T1ce. We choose T1ce for the Student since this is the standard modality used in pre-operative neurosurgery or radiotherapy.

\textbf{Model comparison:} To demonstrate the effectiveness of the proposed framework, we first compare it to a baseline model. Its architecture is the same as the encoder-decoder network in Figure \ref{fig1} and it is trained using only the T1ce modality as input. We also compare it to two other models, U-HVED and HeMIS, using only T1ce as input. Results were directly taken from \cite{shen_hetero-modal_2019}. The results are visible in Table \ref{tab:1}. Our method outperforms U-HVED and HeMIS in the segmentation of all three tumor components. KD-Net also seems to obtain better results than the method proposed in \cite{chen_robust_2019} (again when using only T1ce as input). The authors show results on the BraTS 2015 dataset and therefore they are not directly comparable to KD-Net. Furthermore, we could not find online their code. Nevertheless, the results of HeMIS \cite{havaei_hemis:_2016} on BraTS 2015 (in \cite{chen_robust_2019}) and on BraTS 2018 (in \cite{shen_hetero-modal_2019}) suggest that the observations of BraTS 2018 seem to be more difficult to segment. Since the method proposed in \cite{chen_robust_2019} has worst results than ours on a dataset that seems easier to segment, this should also be the case for the BraTS 2018 dataset. However, this should be confirmed.

\begin{table}[]
    \centering
    \setlength{\tabcolsep}{5pt}
    \renewcommand{\arraystretch}{1.25}
    \caption{Comparison of 3 models using the dice score on the tumor regions. Results of U-HVED and HeMIS are taken from the article \cite{shen_hetero-modal_2019}, where the standard deviations were not provided.}
    \begin{tabular}{l|l l l}
        \hline
         Model & ET & TC & WT  \\
         \hline
         Baseline (nnUnet \cite{isensee_2018}) & $68.1\pm{1.27}$ & $80.28\pm{2.44}$& $\mathbf{77.06\pm{1.47}}$ \\
         Teacher (4 modalities) & $69.47\pm{}1.86$&$	80.77\pm{}1.18$&$88.48\pm{}0.79$ \\
         U-HVED  & $65.5$ & $66.7$ & $62.4$\\
         HeMIS  & $60.8$ & $58.5$ & $58.5$\\
         Ours & $\mathbf{71.67\pm{1.22}}$ & $\mathbf{81.45\pm{1.25}}$ &$ 76.98\pm{1.54}$\\
         \hline
    \end{tabular}
    \label{tab:1}
\end{table}

\textbf{Ablation study:} To evaluate the contribution of each loss term, we did an ablation study by removing each term from the objective function defined in Eq.~\ref{EQobj}. Table \ref{tab:ablation} shows the results using either 0 or 4 skip-connections both in the Student and Teacher networks. We observe that both the KL and KD loss improves the results with respect to the baseline model, especially for the enhanced tumor and tumor core. This also demonstrates that the proposed framework is generic and it works with different encoder-decoder architectures. More results can be found in the supplementary material.

\begin{table}
    \centering
    \setlength{\tabcolsep}{5pt}
    \renewcommand{\arraystretch}{1.25}
    \caption{Ablation study of the loss terms. We compare the results of the model trained with 3 different objective functions: the baseline using only the GT loss, KD-Net trained with only the KL term and KD-Net with the complete objective function. We also tested it with 0 or 4 skip-connections for both the Student and the Teacher.}
    \begin{tabular}{c|l|l||l|l|l}
        \hline
         \begin{tabular}[c]{@{}c@{}}Skip \\ connections\end{tabular} & Model & Loss & ET & TC & WT  \\
         \hline
          4 & Baseline & GT & $68.1\pm{1.27}$ & $80.28\pm{2.44}$& $77.06\pm{1.47}$ \\
          4 & Teacher & GT & $69.47\pm{}1.86$&$	80.77\pm{}1.18$&$88.48\pm{}0.79$ \\
          4 & KD-Net & GT+KL & $70.00\pm{}1.51$&$80.85\pm{}1.82$&$	\mathbf{77.08\pm{1.29}}$\\
          4 & KD-Net & GT+KD & $69.22\pm{}1.19$&$80.54\pm{}1.66$&$	76.83\pm{}1.36$\\
          4 & KD-Net & GT+KL+KD & $\mathbf{71.67\pm1.22}$&$\mathbf{81.45\pm1.25}$&$	76.98\pm1.54$\\
         \hline
         0 & Baseline & GT & $42.95\pm{}3.42$ &$69.44\pm{}1.37$&$	69.41\pm{}1.52$ \\
         0 & Teacher & GT & $42.59\pm{}2.54$&	$69.79\pm{}1.63$&$75.93\pm{}0.33$\\
          0 & KD-Net & GT+KL & $\mathbf{47.59\pm 0.98}$&$7\mathbf{0.96\pm1.73}$&$71.41\pm{}1.2$\\
          0 & KD-Net & GT+KD & $44.8\pm1.1$	& $70.12\pm{}2.42$ & $70.19\pm1.4$\\
          0 & KD-Net & GT+KL+KD & $46.23\pm{}2.91$&$70.73\pm{}2.47$&	$\mathbf{71.93\pm{}1.26}$\\
          \hline
    \end{tabular}
    \label{tab:ablation}
\end{table}

\begin{figure}
\includegraphics[width=\textwidth]{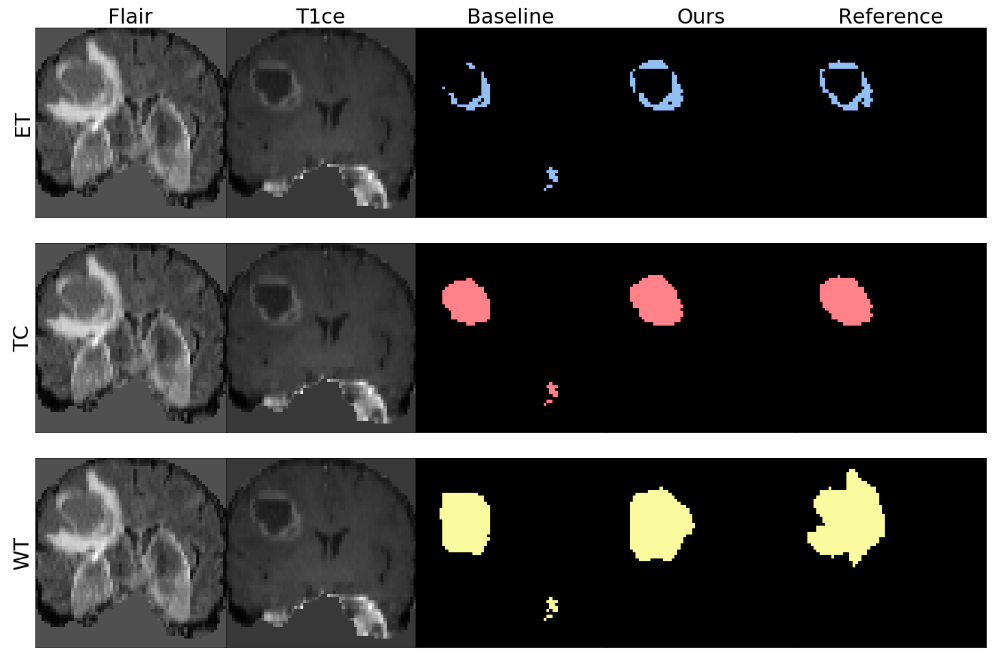}
\caption{Qualitative results obtained using the the baseline and the proposed framework (Student). We show the slice of a subject with the corresponding 3 segmentation labels.} \label{fig2}
\end{figure}

\textbf{Qualitative results:}
In Figure~\ref{fig2}, we show some qualitative results of the proposed framework and compare them with the ones obtained using the baseline method. We can see that the proposed framework allows the Student to discard some outliers and predict segmentation labels of higher quality. In the experiments, the student uses as input only T1ce, which clearly highlights the enhancing tumor. Remarkably, it seems that the Student learns more in this region (see Figure~\ref{fig2} and Table \ref{tab:1}). The knowledge distilled from the Teacher seems to help the Student learn more where it is supposed to be ``stronger". More qualitative results can be found in the supplementary material.

\textbf{Observations:}
It is important to remark that we also tried to expand the Student network by first synthesizing another modality, such as the Flair, from the T1ce and then using it, together with the T1ce, for segmenting the tumor labels. Results were actually worse than the baseline and the computational time quite prohibitive. We also tried sharing the weights between the Teacher and the Student in the deepest layers of the networks to help transferring the knowledge. The intuition behind it was that since the bottlenecks should be the same, the information in the deepest layers should be handled identically. The results were almost identical, but slightly worse, to the ones obtained with the proposed framework presented in Figure~\ref{fig1}. In this paper, we used the nnUNet\cite{isensee_2018} as network for the Student and Teacher, but theoretically any other encoder-decoder architecture, such as the one in \cite{multiResUNet_2020}, could be used. 

\section{Conclusions}
We present a novel framework to transfer knowledge from a multi-modal segmentation network to a mono-modal one. To this end, we propose to use a twofold approach. We employ the strategy of generalized knowledge distillation and, in addition, we also constrain the latent representation of the Student to be similar to the one of the Teacher. We validate our method in brain tumor segmentation, achieving better results than state-of-the-art methods when using only T1ce on Brats 2018. The proposed framework is generic and can be applied to any encoder-decoder segmentation network. The gain in segmentation accuracy and robustness to errors produced by the proposed framework makes it highly valuable for real-world clinical scenarios where only one modality is available at test time.

\section{Acknowledgment}
M.Hu is grateful for financial support from China Scholarship Council.This work is supported by SHEITC (No. 2018-RGZN-02046), 111 plan (No. BP0719010),  and STCSM (No. 18DZ2270700). M. Maillard was supported by a grant of IMT, Fondation Mines-Télécom and Institut Carnot TSN, through the “Futur \& Ruptures” program.

%
%
%
\bibliographystyle{splncs04}
\bibliography{paper2614}
\end{document}